\documentclass[conference]{IEEEtran}
\usepackage{graphicx}
\usepackage{booktabs}
\usepackage{multirow}
\usepackage{amsmath}
\usepackage{xcolor}
\usepackage[hidelinks]{hyperref}
\graphicspath{{figs/}}

\begin{document}

\title{Where Does EEG Help Wearable Freezing of Gait Detection?}


\author{
\IEEEauthorblockN{Chuntian Chi\IEEEauthorrefmark{1},
Lifang He\IEEEauthorrefmark{2},
Huajie Shao\IEEEauthorrefmark{1},
Gang Zhou\IEEEauthorrefmark{1}}
\IEEEauthorblockA{\IEEEauthorrefmark{1}Department of Computer Science, William \& Mary, Williamsburg, VA, USA\\
\IEEEauthorrefmark{2}Department of Computer Science, Lehigh University, Bethlehem, PA, USA\\
Email: \{cchi05, hshao, gzhou\}@wm.edu, lih319@lehigh.edu}
}

\maketitle

\begin{abstract}
Freezing of gait (FoG) detectors built on inertial measurement units (IMUs) are accurate while the legs move, but they become
unreliable when the legs are nearly still, which is exactly when akinetic freezing, voluntary stopping, and standing must be told
apart. Electroencephalography (EEG) captures motor-planning activity that leg sensors cannot, but EEG is impractical for daily
use. We ask where EEG helps: as an input at inference time, or only during pretraining, so that the deployed detector still needs
only IMUs. We assemble a pretraining corpus of synchronized EEG and IMU recordings (5 datasets, 184 people, about 94 hours) and
compare six settings that differ only in the modalities used for pretraining and as model input, each with tuned
hyperparameters and five independent pretraining runs. On the BXHC FoG dataset (leave-one-patient-out, 10 patients with FoG),
adding EEG as input raises FoG F1 from 0.663 to 0.740 (9 of 10 patients improve), and the gain in ranking performance is
confined to the quartile of windows with the least leg movement (AUROC +0.063). Pretraining the IMU encoder jointly with EEG
improves an IMU-only detector both within the cohort (FoG F1 0.663 to 0.705) and when it is trained on a small cohort and tested
on another (FoG F1 +0.083, all 10 test patients improve). Repeating the study with PD-only and healthy-only pretraining corpora,
each with its own hyperparameter search, shows that IMU-based detection is robust to the corpus, whereas the benefit of aligning
EEG with IMU requires both. Cross-cohort analysis further shows that detectors trained without annotated voluntary stops classify
nearly all stops as freezing. Finally, on PD versus non-PD classification in three external datasets, joint EEG+IMU pretraining
improves PD detection from walking EEG over EEG-only pretraining (AUROC 0.750 vs.\ 0.683), while IMU-based detection does not
depend on the pretraining.
\end{abstract}

\section{Introduction}
Freezing of gait is a brief, episodic absence of effective stepping despite the intention to walk~\cite{nutt2011freezing}.
It causes falls and loss of independence in people with Parkinson's disease (PD), and real-time cueing can shorten
episodes, provided that a detector recognizes them reliably~\cite{bachlin2010wearable,Koltermann2024GaitGuardTF}. Most detectors use
IMUs on the legs or trunk~\cite{bachlin2010wearable,mazilu2012online,koltermann2023fogfinder,Koltermann2024GaitGuardTF,borzi2023realtime}.
They work well when freezing changes leg movement, for example the 3--8\,Hz trembling of the legs captured by the freezing
index~\cite{bachlin2010wearable}. They work less well when the legs are nearly still: akinetic freezing, a voluntary stop,
standing, and a slow pivot turn produce similar leg signals, and misclassified turns are a main source of false
alarms~\cite{Koltermann2024GaitGuardTF}.

EEG adds information that leg sensors lack. Freezing is preceded and accompanied by changes in cortical activity~\cite{handojoseno2014analysis},
and EEG-based detectors perform well in the laboratory~\cite{zhang2022multimodal}. However, EEG needs a cap and skin contact,
which makes it impractical for daily monitoring~\cite{Koltermann2024GaitGuardTF}. This raises a practical question: \emph{if EEG is
available only in the laboratory, can it still improve a detector that uses IMUs alone?} One route is to use EEG only during
pretraining, as privileged information~\cite{vapnik2009lupi,gupta2016crossmodal}: the IMU encoder learns representations that
agree with the simultaneous EEG, and the deployed model needs only IMUs.

In this paper, we compare the two routes under controlled conditions. We study six settings that differ only in which
modalities are used for pretraining and which are given to the detector (Table~\ref{tab:xuanwu}), using the same data, folds,
and evaluation for all of them. Our contributions are:
\begin{itemize}
\item A pretraining corpus of synchronized EEG and IMU recordings from five public datasets (184 people, two PD cohorts with
labeled FoG), and a controlled comparison of six pretraining and input settings with tuned hyperparameters and five independent
pretraining runs per setting.
\item We show that EEG as model input improves FoG detection mainly in windows with little leg movement, and that this pattern
holds when the detector is trained on both PD cohorts.
\item We show that EEG used only during pretraining improves an IMU-only detector, most strongly when it is trained on a small
cohort and applied to a different one.
\item We repeat the comparison with PD-only and healthy-only pretraining corpora and show which effects depend on the corpus.
\item We extend the comparison to PD versus non-PD classification on three external datasets with IMU or EEG.
\item We analyze why detectors fail across cohorts: annotated voluntary stops, freezing subtypes, and label conventions differ
between datasets.
\end{itemize}

\section{Data}
\label{sec:data}
\begin{table}[t]
\centering
\caption{Datasets with synchronized EEG and IMU used for pretraining. Only BXHC and SenseFOG have FoG labels.}
\label{tab:data}
\setlength{\tabcolsep}{3pt}
\resizebox{\columnwidth}{!}{%
\begin{tabular}{lllccr}
\toprule
Dataset & Population & People & EEG ch. & IMU sites & Hours \\
\midrule
BXHC~\cite{zhang2022multimodal} & PD & 12 & 25 & shanks, L5$^{a}$ & 3.7 \\
SenseFOG~\cite{klocke2024supraspinal} & PD, STN-DBS & 12 & 48 & ankles, L5 & 2.9 \\
PhysioNet gait~\cite{katmah2026physionet} & healthy, young & 57 & 19 & pelvis, shanks, feet & 2.8 \\
MiM young~\cite{mim_young} & healthy, young & 32 & 120 & back & 27.0 \\
MiM older~\cite{mim_older} & healthy, older & 71 & 120 & back & 57.6 \\
\midrule
Total & & 184 & & & 94.0 \\
\bottomrule
\end{tabular}}\\[2pt]
\footnotesize{$^{a}$L5 sensor available for 6 of 12 patients. MiM: Mind in Motion.}
\end{table}

Table~\ref{tab:data} lists the five datasets. Two contain labeled FoG and are used for evaluation.

\textbf{BXHC}~\cite{zhang2022multimodal}, recorded at Xuanwu Hospital in Beijing, contains 12 patients with PD (mean age 69.1 years, mean disease duration 9.3 years)
recorded off medication during tasks designed to provoke freezing: rising from a chair, walking a narrow corridor with
obstacles, and 180$^\circ$ turns, including turns inside a 0.6\,m square. Two physicians labeled every freezing episode from
video (334 episodes, 88 minutes). The dataset includes 25-channel EEG, released after artifact removal with independent
component analysis, and IMUs on both shanks, the lower back (L5), and the arm. Following the dataset authors, we evaluate on the
10 patients who froze during the recordings.

\textbf{SenseFOG}~\cite{klocke2024supraspinal} contains 12 patients with PD and bilateral subthalamic deep brain stimulation
(mean disease duration 12.8 years), recorded after overnight medication withdrawal with stimulation switched off 20 minutes
before the session. Patients walked a hallway narrowed to provoke freezing, including a condition with instructed voluntary
stops. Freezing labels cover straight-line walking only (episodes of at least 1\,s; freezing during turns and gait initiation
was excluded), and almost all labeled freezes are of the trembling-in-place type. IMUs were worn at both ankles and L5, together
with 48-channel EEG. Freezing occurred in 5 patients; we report FoG metrics on the 4 patients with at least 10 freezing windows.
Because labels exist only for annotated segments, we use only windows that lie entirely inside a labeled segment.

The three healthy datasets~\cite{katmah2026physionet,mim_young,mim_older} contribute unlabeled walking at several speeds and
on uneven terrain. All IMU data are expressed in a body-fixed frame (acceleration in $g$ including gravity, angular velocity in
rad/s) and resampled to 100\,Hz. BXHC EEG is used as released; all other EEG is high-pass filtered at 0.5\,Hz and
notch-filtered at the local mains frequency. All EEG is resampled to 200\,Hz and z-scored per channel and recording. We cut all recordings into 1.28\,s windows (stride 0.64\,s), the
window length used in prior real-time FoG detection~\cite{Koltermann2024GaitGuardTF}. A window is labeled FoG if freezing occurs in its last
30\%~\cite{Koltermann2024GaitGuardTF}.

\section{Method}
\label{sec:method}
\subsection{Encoders}
The \textbf{IMU encoder} is a masked autoencoder~\cite{he2022mae} (5.2M parameters; 0.1\,s patches; 6 transformer layers)
previously pretrained on 1,682 hours of IMU data from 11 public datasets (Appendix~\ref{app:imu}). It accepts any set of
sensors and encodes each sensor's body location. The \textbf{EEG encoder} is REVE~\cite{reve2025}, a pretrained EEG foundation
model that accepts arbitrary electrode montages through electrode positions; we use its token for the most recent 1\,s of each
window.

\subsection{Pretraining}
Starting from these two encoders, we continue pretraining on the corpus in Table~\ref{tab:data} (all windows, no FoG labels).
Each training batch contains windows from one recording, and half of the batches come from the two PD cohorts.
\begin{itemize}
\item \textbf{IMU pretraining} continues the IMU encoder with its masked reconstruction loss.
\item \textbf{EEG pretraining} continues the top transformer blocks of the EEG encoder with masked reconstruction of
whole channels.
\item \textbf{EEG+IMU pretraining} continues one encoder with its own reconstruction loss plus a contrastive
loss~\cite{oord2018cpc,radford2021clip} that matches its representation of a 1\,s segment to the representation of the
simultaneous 1\,s segment from the other modality, computed by that modality's frozen encoder. Negatives are other moments of
the same recording at least 1\,s apart. Applied to the IMU encoder, this yields an IMU encoder whose representation is shaped by
EEG; applied to the EEG encoder, it yields an EEG encoder shaped by IMU.
\end{itemize}
We refer to a setting by the modalities used in pretraining and the modalities given to the detector, written
\emph{pretraining $\rightarrow$ input}. For example, ``EEG+IMU $\rightarrow$ IMU'' uses the IMU encoder from EEG+IMU
pretraining and receives only IMU data, both during training and at test time.

\subsection{Settings and detection model}
Table~\ref{tab:xuanwu} lists the six settings. S1, S2, and S6 form the comparisons of interest: S2$-$S1 isolates the effect of
EEG during pretraining with IMU-only input, and S6$-$S2 isolates the effect of adding EEG as input. S5 combines the separately
pretrained encoders, and R1 and R2 are the encoders before continued pretraining. Encoders are frozen. For each window, a small
head is trained on the encoder outputs: a two-layer perceptron on the pooled IMU embedding, attention pooling over EEG channels
followed by a perceptron, and, for EEG+IMU input, a gated fusion of both that randomly drops the EEG input during training.

\subsection{Evaluation}
We use leave-one-patient-out cross-validation. In each fold, two other patients serve as inner validation for early stopping
and for choosing the decision threshold (the threshold that maximizes mean FoG F1 over these patients). All hyperparameters
(learning rate, training length, share of PD batches, and replay of the original IMU corpus for IMU pretraining; trainable
blocks and mask ratio for EEG pretraining; contrastive weight, temperature, and whether alignment uses all data or only PD
data; and the head) were selected by mean inner-validation AUPRC on BXHC, with the same budget for every setting
(Appendix~\ref{app:tuning}, Fig.~\ref{fig:tuning}). Test patients were never used for selection, and SenseFOG was never used for selection. Each
selected setting was then pretrained five times with independent random seeds. We report the mean and standard deviation (SD)
across patients of replicate-averaged results. Paired comparisons use per-patient differences with 95\% bootstrap confidence
intervals and two-sided Wilcoxon signed-rank tests. Pretraining used unlabeled windows of all patients, including the test
patient of each fold; labels of the test patient were never used.

To locate where EEG helps, we split windows into quartiles by leg movement, measured as the standard deviation of the
acceleration magnitude within the window (cut-points computed per cohort), and compute AUROC within each quartile.

\section{Results}
\label{sec:results}
\begin{table*}[t]
\centering
\caption{BXHC, leave-one-patient-out (10 patients with FoG). Mean $\pm$ SD across patients; across the five pretraining
runs, the SD of every metric is at most 0.033. Best result per column in bold (ties both in bold).}
\label{tab:xuanwu}
\resizebox{\textwidth}{!}{%
\begin{tabular}{llllccccc}
\toprule
 & Pretraining & Input & Encoder(s) & FoG F1 & Weighted F1 & Sensitivity & Specificity & AUROC\\
\midrule
R1 & none (base) & IMU & base IMU encoder & 0.622$\pm$0.271 & 0.766$\pm$0.125 & 0.640$\pm$0.354 & 0.800$\pm$0.190 & 0.897$\pm$0.053\\
S1 & IMU & IMU & IMU-pretrained & 0.663$\pm$0.235 & 0.786$\pm$0.108 & 0.675$\pm$0.311 & 0.812$\pm$0.165 & 0.906$\pm$0.050\\
S2 & EEG+IMU & IMU & EEG+IMU-pretrained IMU & 0.705$\pm$0.207 & 0.809$\pm$0.106 & 0.730$\pm$0.288 & 0.835$\pm$0.136 & \textbf{0.912$\pm$0.047}\\
\midrule
R2 & none (base) & EEG & general REVE & 0.533$\pm$0.191 & 0.377$\pm$0.191 & \textbf{0.865$\pm$0.234} & 0.139$\pm$0.227 & 0.574$\pm$0.057\\
S3 & EEG & EEG & EEG-pretrained & 0.519$\pm$0.212 & 0.395$\pm$0.195 & 0.831$\pm$0.287 & 0.174$\pm$0.273 & 0.575$\pm$0.076\\
S4 & EEG+IMU & EEG & EEG+IMU-pretrained EEG & 0.575$\pm$0.166 & 0.523$\pm$0.172 & 0.829$\pm$0.183 & 0.341$\pm$0.247 & 0.666$\pm$0.101 \\
\midrule
S5 & IMU; EEG (separate) & EEG+IMU & S1 + S3 encoders & 0.725$\pm$0.158 & 0.812$\pm$0.065 & 0.754$\pm$0.213 & 0.820$\pm$0.121 & 0.905$\pm$0.047 \\
S6 & EEG+IMU & EEG+IMU & S2 + S4 encoders & \textbf{0.740$\pm$0.177} & \textbf{0.830$\pm$0.050} & 0.763$\pm$0.218 & \textbf{0.842$\pm$0.092} & \textbf{0.912$\pm$0.044} \\
\bottomrule
\end{tabular}}
\end{table*}

\begin{figure*}[t]
\centering
\includegraphics[width=\textwidth]{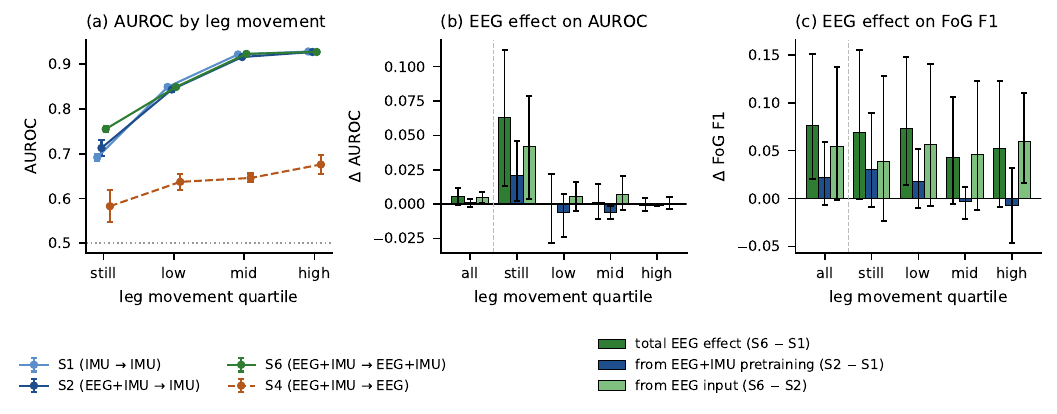}
\vspace{-2.5em}
\caption{BXHC by leg-movement quartile (``still'' = least movement). (a) AUROC per quartile, mean $\pm$ SD across the five
pretraining runs. (b, c) The total effect of EEG (S6$-$S1) split into the part from EEG+IMU pretraining (S2$-$S1) and the part
from EEG input (S6$-$S2); mean over patients with 95\% bootstrap confidence intervals.}
\vspace{-1em}
\label{fig:quartiles}
\end{figure*}

\subsection{EEG as input improves detection on BXHC}
Table~\ref{tab:xuanwu} shows the results on BXHC. With IMU input, continued pretraining on the new corpus (S1) improves
over the base encoder (R1): FoG F1 rises from 0.622 to 0.663 (7 of 10 patients improve; $p=0.049$) and AUROC from 0.897 to
0.906 (9 of 10; $p=0.004$). Adding EEG as input raises FoG F1 to 0.740 (S6 vs.\ S1: $+$0.077, 95\% CI [$+$0.022, $+$0.149],
9 of 10 patients improve, $p=0.020$). Both sensitivity and specificity increase, and the SD across patients falls from 0.235 to
0.177, so the gain is largest for patients on whom the IMU-only detector performs worst. AUROC improves only slightly ($+$0.006;
9 of 10; $p=0.037$), which indicates that EEG mainly helps the decision threshold transfer to an unseen patient. EEG alone
performs close to chance (S3, AUROC 0.575). Per-patient results and the variation across pretraining runs are shown in
Appendix~\ref{app:patients} (Fig.~\ref{fig:main}).

EEG+IMU pretraining of the IMU encoder also improves IMU-only detection (S2 vs.\ S1): FoG F1 rises from 0.663 to 0.705,
AUROC from 0.906 to 0.912, and both sensitivity (0.675 to 0.730) and specificity (0.812 to
0.835) increase. The difference between S6 and S5, which uses the same inputs with separately pretrained encoders,
is also small (FoG F1 $+$0.015, $p=0.13$).

\subsection{EEG helps when the legs are still}
Figure~\ref{fig:quartiles}a shows that all IMU-based settings perform well when the legs move (AUROC 0.85--0.93) and poorly
in the stillest quartile (S1: 0.692). This is the only quartile in which EEG input changes the ranking: S6 improves the stillest
quartile by $+$0.063 (95\% CI [$+$0.014, $+$0.112]; 9 of 10 patients; $p=0.037$), while the other three quartiles change by
less than 0.002 (Figure~\ref{fig:quartiles}b). About one third of this gain comes from EEG+IMU pretraining and two thirds from
EEG input. For FoG F1, the gain is spread across quartiles (Figure~\ref{fig:quartiles}c), consistent with a better decision
threshold.

The EEG encoder shaped by IMU (S4) shows the opposite pattern. It improves EEG-only detection overall (AUROC 0.575 to 0.666;
9 of 10 patients; $p=0.004$), but the gains appear in the three quartiles with leg movement ($+$0.08 to $+$0.10; 10 of 10
patients for the low and mid quartiles; $p=0.002$) and not in the stillest quartile ($+$0.03; $p=0.49$). The same
pattern holds in all five evaluation settings (Appendix~\ref{app:s4}, Fig.~\ref{fig:s4}). EEG recorded during walking contains movement-related signals~\cite{gwin2010removal}, and aligning
EEG with IMU appears to teach the EEG encoder to read them. The benefit of EEG as an additional input, in contrast, appears only
where leg movement and therefore movement-related EEG signals are smallest.

\subsection{Generalization across cohorts}
\begin{table*}[t]
\centering
\caption{Generalization. SenseFOG: leave-one-patient-out within SenseFOG (4 patients with FoG). Combined: leave-one-patient-out
over both cohorts (14 patients with FoG). Transfer: trained on one cohort, threshold chosen on that cohort, tested on all
patients of the other. Mean $\pm$ SD across patients; best per column in bold.}
\label{tab:general}
\resizebox{\textwidth}{!}{%
\begin{tabular}{lll cc cc cc cc}
\toprule
 & & & \multicolumn{2}{c}{SenseFOG} & \multicolumn{2}{c}{Combined} & \multicolumn{2}{c}{BXHC$\rightarrow$SenseFOG} & \multicolumn{2}{c}{SenseFOG$\rightarrow$BXHC} \\
 & Pretraining & Input & FoG F1 & AUROC & FoG F1 & AUROC & FoG F1 & AUROC & FoG F1 & AUROC \\
\midrule
R1 & none & IMU & 0.735$\pm$0.295 & 0.924$\pm$0.042 & 0.645$\pm$0.256 & 0.918$\pm$0.058 & 0.615$\pm$0.170 & \textbf{0.974$\pm$0.020} & 0.630$\pm$0.190 & 0.856$\pm$0.035 \\
S1 & IMU & IMU & 0.746$\pm$0.260 & 0.902$\pm$0.076 & 0.693$\pm$0.227 & 0.926$\pm$0.052 & 0.733$\pm$0.131 & 0.973$\pm$0.018 & 0.648$\pm$0.168 & 0.843$\pm$0.051 \\
S2 & EEG+IMU & IMU & 0.730$\pm$0.285 & 0.945$\pm$0.047 & 0.694$\pm$0.227 & 0.924$\pm$0.053 & 0.685$\pm$0.148 & 0.971$\pm$0.018 & \textbf{0.731$\pm$0.155} & \textbf{0.891$\pm$0.042} \\
\midrule
R2 & none & EEG & 0.453$\pm$0.180 & 0.489$\pm$0.073 & 0.516$\pm$0.191 & 0.565$\pm$0.077 & 0.416$\pm$0.173 & 0.498$\pm$0.089 & 0.444$\pm$0.118 & 0.506$\pm$0.032 \\
S3 & EEG & EEG & 0.461$\pm$0.189 & 0.459$\pm$0.050 & 0.525$\pm$0.196 & 0.555$\pm$0.068 & 0.437$\pm$0.201 & 0.526$\pm$0.092 & 0.265$\pm$0.070 & 0.485$\pm$0.024 \\
S4 & EEG+IMU & EEG & 0.461$\pm$0.199 & 0.553$\pm$0.049 & 0.539$\pm$0.177 & 0.646$\pm$0.094 & 0.364$\pm$0.173 & 0.575$\pm$0.086 & 0.554$\pm$0.141 & 0.641$\pm$0.094 \\
\midrule
S5 & separate & EEG+IMU & \textbf{0.770$\pm$0.217} & 0.970$\pm$0.019 & 0.705$\pm$0.180 & 0.927$\pm$0.052 & \textbf{0.810$\pm$0.117} & 0.964$\pm$0.021 & 0.702$\pm$0.151 & 0.878$\pm$0.050 \\
S6 & EEG+IMU & EEG+IMU & 0.757$\pm$0.171 & \textbf{0.974$\pm$0.017} & \textbf{0.732$\pm$0.164} & \textbf{0.930$\pm$0.048} & 0.749$\pm$0.154 & 0.960$\pm$0.023 & 0.705$\pm$0.166 & 0.885$\pm$0.043 \\
\bottomrule
\end{tabular}}
\end{table*}

\begin{figure*}[t]
\centering
\includegraphics[width=\textwidth]{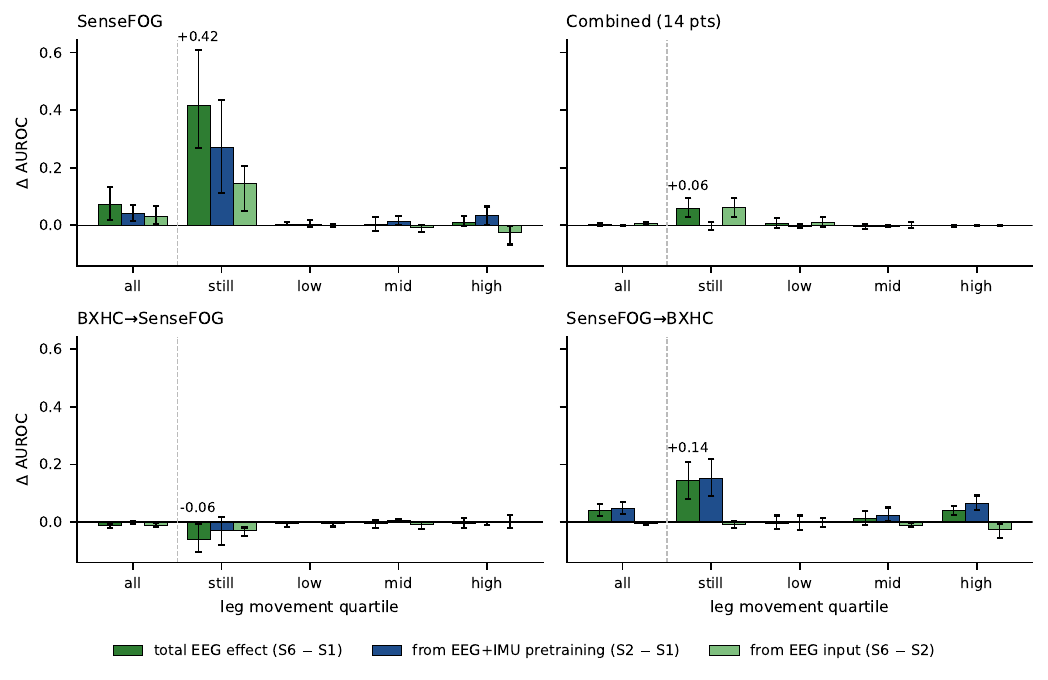}
\caption{The AUROC decomposition of Fig.~\ref{fig:quartiles}b in the four other evaluation settings (shared y-axis; the total
effect in the stillest quartile is printed above each panel's bar). Mean over patients with 95\% bootstrap confidence intervals.}
\label{fig:decomp}
\end{figure*}

Table~\ref{tab:general} and Figure~\ref{fig:decomp} repeat the comparison on SenseFOG, on both cohorts combined, and across
cohorts, with the settings and hyperparameters selected on BXHC.

\textbf{SenseFOG and combined training.} Within SenseFOG, IMU input alone is already strong (S1 AUROC 0.902), and the stillest
quartile is again the weak spot (AUROC 0.457). EEG input raises it by 0.42 in all 4 patients with FoG. In the combined
setting, with both cohorts in training, the IMU-only detector improves on SenseFOG (stillest-quartile AUROC 0.839), and the
gain from EEG input becomes smaller. Pooled over all 14 patients, EEG input still improves the stillest quartile ($+$0.059;
95\% CI [$+$0.030, $+$0.092]; 11 of 14 patients; $p=0.003$) and leaves the other quartiles unchanged ($|\Delta| < 0.01$).
The large gain within SenseFOG alone therefore partly compensates for the small number of patients with FoG available for
training.

\textbf{Transfer from SenseFOG to BXHC.} Here the detector is trained on SenseFOG alone and tested on BXHC. With IMU input,
the IMU encoder from EEG+IMU pretraining outperforms the one from IMU pretraining in every test patient: FoG F1 0.731 vs.\ 0.648
($+$0.083; 95\% CI [$+$0.055, $+$0.114]; 10 of 10; $p=0.002$) and AUROC 0.891 vs.\ 0.843 (9 of 10; $p=0.006$). The gain is
concentrated in the stillest quartile ($+$0.152; 10 of 10; $p=0.002$), and results are far more stable across pretraining runs
(FoG F1 SD 0.004 vs.\ 0.110). SenseFOG's labeled freezes are almost all of the trembling-in-place type, while BXHC also
contains freezing during turns and gait initiation, where the legs are nearly still. EEG+IMU pretraining thus helps exactly
where the training data lack examples. Adding EEG as input in this direction adds little beyond S2 (S6$-$S2 near zero;
Fig.~\ref{fig:decomp}).

\textbf{Transfer from BXHC to SenseFOG.} Trained on BXHC, the IMU-only detector ranks SenseFOG windows well (AUROC 0.973),
but its specificity drops to 0.61. The false alarms come mostly from voluntary stops: 99.7\% of SenseFOG's stop windows are
classified as freezing, compared with 32\% of its walking windows. BXHC contains no annotated voluntary stops, while its
freezing labels include episodes during turns and gait initiation with little leg movement, so the detector learns to
associate still legs with freezing. With a threshold chosen per patient, the same model reaches FoG F1 0.89, so the
representation transfers and the threshold does not. EEG input reduces false alarms during walking to the in-cohort level
(15\%), but not during stops (84\%), and EEG does not improve ranking in this direction (Fig.~\ref{fig:decomp}). Within
SenseFOG, settings that use EEG during pretraining or as input also separate freezing from voluntary stops better than IMU
pretraining alone (pooled AUROC 0.70--0.76 vs.\ 0.42; Appendix~\ref{app:stops}, Fig.~\ref{fig:stops}).

\subsection{Pretraining corpus}
\label{sec:corpus}
\begin{figure*}[t]
\centering
\includegraphics[width=\textwidth]{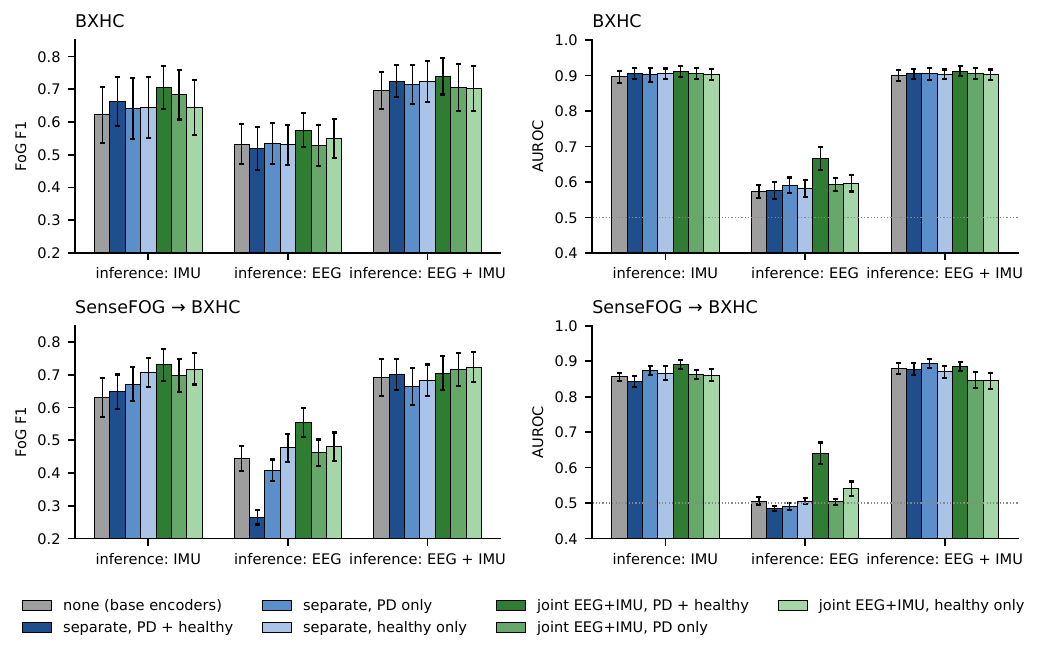}
\caption{FoG F1 and AUROC by pretraining, pretraining corpus and inference input (BXHC: leave-one-patient-out; SenseFOG$\rightarrow$BXHC:
trained on SenseFOG, tested on BXHC). Bars: mean across the 10 BXHC patients with FoG (five pretraining runs; base encoders: five
head seeds); error bars: standard error across patients. Every corpus has its own hyperparameter search.}
\label{fig:corpus}
\end{figure*}

To test what the pretraining data contribute, we repeated the procedure, including the full hyperparameter search, with two
reduced corpora: PD data only (BXHC and SenseFOG) and healthy data only (PhysioNet and Mind in Motion). The search omitted
options without meaning within one cohort (the share of PD batches, the alignment data, and replay of the original IMU corpus).
Figure~\ref{fig:corpus} shows every combination of pretraining, corpus, and inference input; the gray bars include the base
encoders with EEG+IMU input, which needs no continued pretraining.

Three results stand out. First, EEG at inference helps without any continued pretraining: fusing the two base encoders raises
FoG F1 on BXHC from 0.622 to 0.696 and stillest-quartile AUROC from 0.657 to 0.719. Second, on BXHC, settings with IMU input
are insensitive to the corpus (AUROC within 0.01), whereas the EEG encoder shaped by IMU needs the full corpus: its AUROC falls
from 0.666 to 0.593 (PD only) and 0.596 (healthy only), with all 10 patients lower in both cases ($p=0.002$), to the level of
EEG-only pretraining. Third, under transfer from SenseFOG, joint EEG+IMU pretraining on the full corpus gives the best IMU-only
detector (FoG F1 0.731, AUROC 0.891). IMU-only pretraining on healthy data comes close (FoG F1 0.708, AUROC 0.866), and the
jointly pretrained encoder keeps a small but consistent AUROC advantage over it ($+$0.026; 95\% CI [$+$0.006, $+$0.045];
7 of 10; $p=0.049$). With the reduced corpora, adding EEG to pretraining does not improve the IMU-only detector under transfer.

\section{PD versus Non-PD Classification}
\label{sec:pdcls}
\begin{table*}[t]
\centering
\caption{PD versus non-PD, subject-level AUROC (mean $\pm$ SD over five runs; 5-fold cross-validation by subject; heads tuned per
setting). Every column is a dataset in which both groups were recorded at the same site with the same protocol. EEG+IMU inference
is not possible: no dataset provides both. Age-matched: AUROC of the run-averaged predictions in an age-matched subset. Best per
column in bold.}
\label{tab:pdcls}
\resizebox{\textwidth}{!}{%
\begin{tabular}{ll ccc|cc}
\toprule
 & & \multicolumn{3}{c|}{IMU inference (R1 / S1 / S2)} & \multicolumn{2}{c}{EEG inference (R2 / S3 / S4)} \\
Pretraining & Corpus & WearGait-PD & PADS: PD vs.\ healthy & PADS: PD vs.\ other & PD-EEG, walking & PD-EEG, rest \\
\midrule
none (R1 / R2) & -- & 0.725$\pm$0.013 & 0.763$\pm$0.013 & \textbf{0.632$\pm$0.005} & 0.752$\pm$0.032 & 0.711$\pm$0.017 \\
\midrule
separate (S1 / S3) & PD + healthy & 0.724$\pm$0.014 & 0.749$\pm$0.009 & 0.621$\pm$0.017 & 0.683$\pm$0.021 & 0.710$\pm$0.029 \\
 & PD only & 0.728$\pm$0.015 & 0.752$\pm$0.011 & 0.613$\pm$0.014 & 0.782$\pm$0.017 & \textbf{0.719$\pm$0.024} \\
 & healthy only & 0.720$\pm$0.013 & 0.771$\pm$0.017 & 0.630$\pm$0.014 & 0.647$\pm$0.006 & 0.678$\pm$0.020 \\
\midrule
joint (S2 / S4) & PD + healthy & 0.731$\pm$0.011 & 0.761$\pm$0.011 & 0.613$\pm$0.020 & 0.750$\pm$0.026 & 0.703$\pm$0.017 \\
 & PD only & \textbf{0.735$\pm$0.020} & 0.750$\pm$0.020 & 0.614$\pm$0.009 & \textbf{0.791$\pm$0.019} & 0.710$\pm$0.018 \\
 & healthy only & 0.729$\pm$0.012 & \textbf{0.772$\pm$0.013} & 0.619$\pm$0.015 & 0.764$\pm$0.038 & 0.699$\pm$0.030 \\
\midrule
age only (baseline) & & 0.730 & 0.558 & 0.560 & 0.649 & 0.688 \\
best setting (bold), age-matched subset & & 0.709 & 0.806 & 0.614 & 0.815 & 0.557 \\
age only, age-matched subset & & 0.484 & 0.497 & 0.486 & 0.478 & 0.497 \\
\bottomrule
\end{tabular}}
\end{table*}

To test whether the encoders support tasks beyond FoG, we classified people with PD versus people without PD in three external
datasets, none of which was used for pretraining. We used only datasets in which both groups were recorded at the same site with
the same protocol, because pooling a PD cohort with a separately recorded control cohort lets a model separate the datasets
instead of the groups. No public dataset provides synchronized EEG and IMU for both groups, so each dataset tests one inference
modality. \textbf{WearGait-PD}~\cite{weargait2024} (version 1, cross-sectional) contains 100 people with PD and 85 controls with
13 body-worn IMUs; we used the self-paced and hurried walks and the sensors at the lower back, both shanks, and both
ankles. \textbf{PADS}~\cite{pads2024} contains wrist IMUs from 276 people with PD, 79 healthy people, and 114 people with other
movement disorders during 11 seated tasks; we report PD versus healthy and PD versus other disorders. \textbf{PD-EEG}~\cite{pdeeg2026}
contains 60-channel EEG from 116 people with PD and 28 controls, recorded during treadmill walking (133 people) and at rest.
Preprocessing matches the pretraining data.

The label belongs to the person, so the evaluation is at the subject level: 5-fold cross-validation by subject, stratified by
group and identical for all settings, with 20\% of each training fold held out for early stopping and the decision threshold.
A small head pools a segment of consecutive window embeddings with attention, and the subject's probability is the mean over its
segments. For every setting, 16 head configurations, including the segment length (4, 6, or 10\,s), were searched on the inner
validation sets; the selected head was then trained with each of the five pretraining runs (base encoders: five seeds). Because
age differs between groups in all three datasets (WearGait-PD: controls 74.1 vs.\ PD 67.0 years; PD-EEG: PD 66.1 vs.\ controls
60.8), we report an age-only baseline (logistic regression on age, same folds) and AUROC in an age-matched subset (nearest-age
pairs within 3 years).

Table~\ref{tab:pdcls} shows the results. With IMU inference, all settings perform alike (WearGait-PD 0.720--0.735,
PADS PD vs.\ healthy 0.749--0.772), and continued pretraining does not improve over the base encoder. In WearGait-PD, age alone
reaches the same AUROC (0.730), but the models keep their AUROC in the age-matched subset (0.70--0.73) while age alone falls to
chance, so the models capture group differences beyond age. Separating PD from other movement disorders is harder (AUROC
0.61--0.63). Walking EEG separates the groups best, and here pretraining matters: joint EEG+IMU pretraining outperforms EEG-only
pretraining (S4 vs.\ S3: 0.750 vs.\ 0.683; $\Delta$AUROC $+$0.082, 95\% bootstrap CI [$+$0.004, $+$0.169]; with the healthy-only
corpus, $+$0.132 [$+$0.037, $+$0.236]), mirroring its effect on EEG-only FoG detection. EEG-only pretraining on the full corpus
lowers AUROC below the base REVE encoder (0.683 vs.\ 0.752), and PD-only pretraining gives the best results (S4: 0.791, S3: 0.782),
which hold in the age-matched subset (0.815--0.820). Resting EEG separates the groups mainly through age: AUROC is close to the
age-only baseline and drops to 0.52--0.61 in the age-matched subset. The control groups are small (PD-EEG: 23 controls with
walking EEG), so these differences are indicative.

\section{Discussion and Limitations}
EEG helped in two different ways depending on the situation. When labeled data from the same cohort were available, EEG as an
additional input improved detection, and its effect on ranking was confined to windows with little leg movement. When the
detector had to be trained on a small cohort and applied to another, EEG used only during pretraining produced an IMU-only
detector that generalized better, mainly in the same still windows. Both results point to the period in which leg signals are
least informative, which is also where false alarms from stops and turns arise~\cite{Koltermann2024GaitGuardTF}.

Our study has several limitations. The labeled cohorts are small (10 and 4 patients with FoG), so the SenseFOG results show
direction rather than significance; with 4 patients, the smallest possible Wilcoxon $p$ is 0.125. The analysis of leg-movement
quartiles was motivated by preliminary analyses on BXHC; SenseFOG and the combined setting provide independent support. Whether the gain from EEG input in still windows reflects cortical activity, muscle activity, or residual motion requires artifact-controlled analyses. Finally, the cohorts differ in freezing subtypes, voluntary stops, medication and stimulation state, and EEG preprocessing, so cross-cohort results reflect all of these differences at once. We suggest that FoG datasets report freezing subtypes, task context, and the handling of voluntary stops along with sensor placement.

\section{Related Work}
IMU-based FoG detection ranges from frequency thresholds~\cite{bachlin2010wearable} and feature-based classifiers~\cite{mazilu2012online}
to neural networks deployed on phones~\cite{koltermann2023fogfinder,Koltermann2024GaitGuardTF} and single-sensor models~\cite{borzi2023realtime};
reducing false alarms during turns and stops remains a central challenge~\cite{Koltermann2024GaitGuardTF}. EEG features predict and detect
freezing~\cite{handojoseno2014analysis}, and combining EEG with IMU, EMG, and skin conductance improves laboratory
detection~\cite{zhang2022multimodal}. Using a modality only during training has been studied as learning using privileged
information~\cite{vapnik2009lupi} and cross-modal distillation~\cite{gupta2016crossmodal}; our EEG+IMU pretraining follows the
contrastive alignment used in multimodal pretraining~\cite{oord2018cpc,radford2021clip}, applied to time-matched physiological
signals.

\section{Conclusion}
We compared EEG as model input and EEG used only in pretraining for wearable FoG detection, under identical data, folds, and
tuning. EEG as input improved FoG F1 on BXHC from 0.663 to 0.740, with its ranking gain confined to windows with little leg
movement, a pattern that held when training on both cohorts and without continued pretraining. EEG used only in pretraining
improved an IMU-only detector within the cohort (FoG F1 0.663 to 0.705) and, most clearly, when it was trained on one cohort
and tested on another (FoG F1 $+$0.083 in all 10 test patients). With PD-only or healthy-only pretraining corpora, IMU-based
detection was unchanged, whereas the benefit of aligning EEG with IMU required both. In PD versus non-PD classification, joint
EEG+IMU pretraining improved PD detection from walking EEG over EEG-only pretraining. These results suggest that laboratory EEG
can improve deployable IMU-only detectors when labeled data are scarce, and that FoG benchmarks should account for voluntary
stops and freezing subtypes.

\bibliographystyle{IEEEtran}
\bibliography{refs}

\appendix
\subsection{IMU pretraining corpus}
\label{app:imu}
The base IMU encoder (R1) was pretrained on 1,682 hours from 11 public IMU datasets (309 people; Table~\ref{tab:imucorpus}).

\begin{table}[h]
\centering
\caption{IMU-only corpus used to pretrain the base IMU encoder (R1).}
\label{tab:imucorpus}
\setlength{\tabcolsep}{3pt}
\resizebox{\columnwidth}{!}{%
\begin{tabular}{llrlrr}
\toprule
Dataset & Group & Subjects & Sensor locations used & Hz & Hours \\
\midrule
DeFoG, daily living~\cite{salomon2024contest} & PD & 65 & lower back & 100 & 1,560.0 \\
DeFoG~\cite{salomon2024contest} & PD & 45 & lower back & 100 & 66.8 \\
tDCS-FoG~\cite{salomon2024contest} & PD & 62 & lower back & 128 & 15.1 \\
PAMAP2~\cite{reiss2012pamap2} & healthy & 9 & wrist, chest, ankle & 100 & 10.7 \\
MotionSense~\cite{malekzadeh2019motionsense} & healthy & 24 & front pocket (phone) & 50 & 7.7 \\
HAPT~\cite{reyesortiz2016hapt} & healthy & 30 & waist (phone) & 50 & 6.2 \\
Daphnet~\cite{bachlin2010wearable} & PD & 10 & shank, thigh, lower back & 64 & 4.9 \\
VCU~\cite{koltermann2023fogfinder} & PD & 23 & left and right ankle & 100 & 3.9 \\
BXHC, unlabeled~\cite{zhang2022multimodal} & PD & 12 & left and right shank & 100 & 3.4 \\
FoG-STAR~\cite{fogstar} & PD & 22 & ankles, lower back, wrist & 60 & 1.5 \\
Stanford IMU-FoG~\cite{oday2022assessing} & PD & 7 & 11 body sites & 128 & 1.4 \\
\midrule
Total & & 309 & & & 1,681.6 \\
\bottomrule
\end{tabular}}
\end{table}

\subsection{Hyperparameter search}
\label{app:tuning}
\begin{figure}[!h]
\centering
\includegraphics[width=\columnwidth]{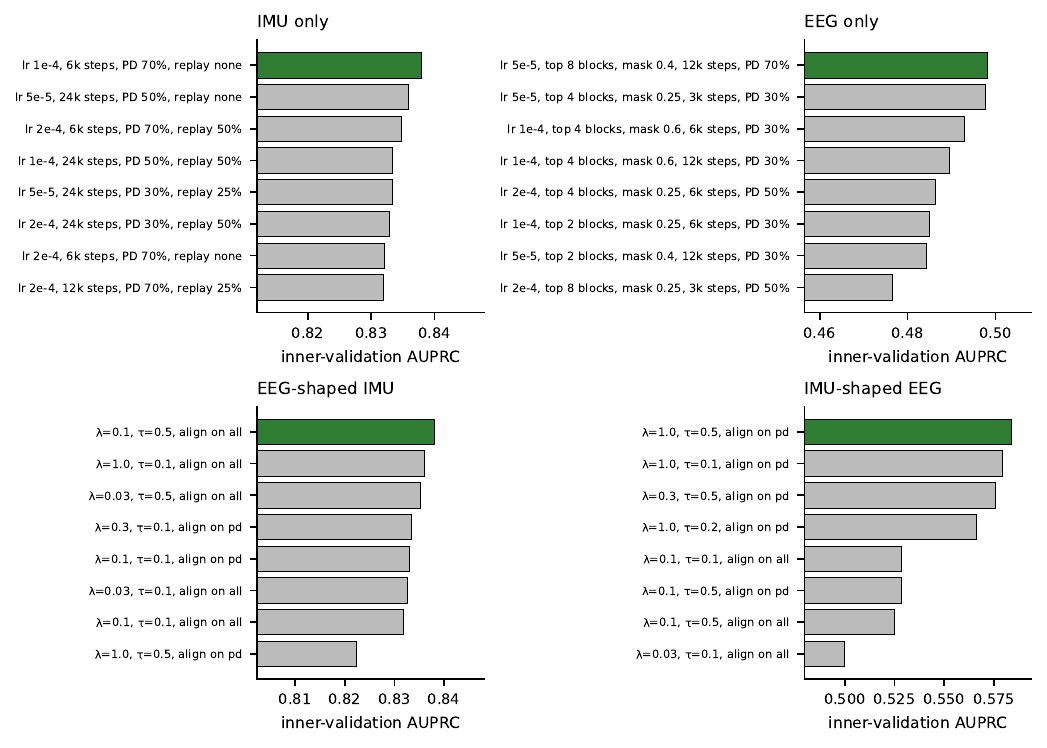}
\caption{Hyperparameter search: inner-validation AUPRC on BXHC of each configuration (selected in green).}
\label{fig:tuning}
\end{figure}

Each encoder family was given eight random configurations, and each head sixteen, all scored by mean inner-validation AUPRC
over the 12 BXHC folds. Figure~\ref{fig:tuning} shows the score of every configuration. The IMU encoder was insensitive to
its settings: all IMU pretraining configurations scored between 0.832 and 0.838, and all EEG+IMU configurations of the IMU
encoder between 0.822 and 0.838. The EEG encoder shaped by IMU was the exception: configurations that aligned on PD data only
with a contrastive weight of 0.3--1.0 scored 0.57--0.58, while configurations that aligned on all data scored 0.50--0.53,
close to EEG pretraining alone (0.50). Selected settings: IMU pretraining, learning rate $10^{-4}$, 6k steps, 70\% PD batches,
no replay; EEG pretraining, learning rate $5\times10^{-5}$, top 8 blocks trainable, mask ratio 0.4, 12k steps; EEG+IMU
pretraining of the IMU encoder, contrastive weight 0.1, temperature 0.5, alignment on all data; EEG+IMU pretraining of the EEG
encoder, weight 1.0, temperature 0.5, alignment on PD data only.

\subsection{Per-patient results on BXHC}
\label{app:patients}
\begin{figure}[!h]
\centering
\includegraphics[width=\columnwidth]{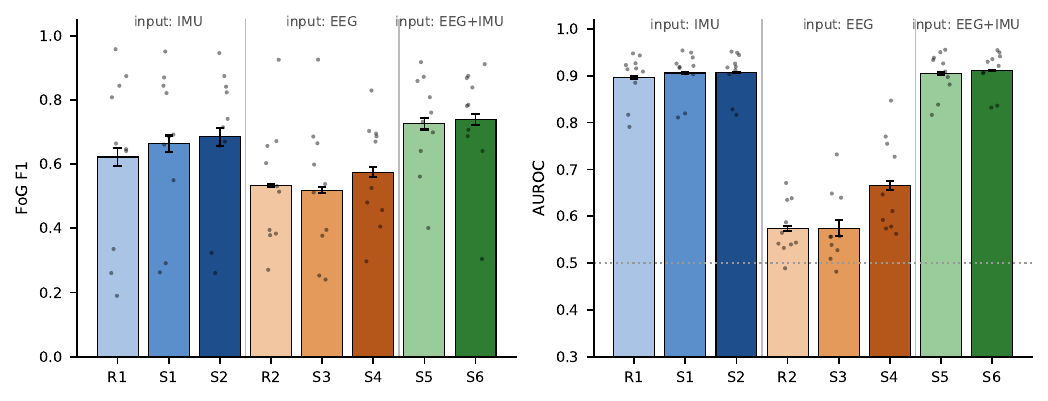}
\caption{BXHC results per setting. Bars: mean over five pretraining runs; error bars: SD across runs; dots: patients.}
\label{fig:main}
\end{figure}

Figure~\ref{fig:main} shows FoG F1 and AUROC for every setting on BXHC, with the mean over the five pretraining runs, the
SD across runs, and each patient's replicate-averaged result. Run-to-run variation is small (SD at most 0.033 for every metric)
compared with the variation between patients (SD 0.16--0.27 for FoG F1), so differences between settings are not driven by
individual pretraining runs. With EEG+IMU input, the between-patient SD of FoG F1 decreases from 0.235 (S1) to 0.177 (S6).
All three EEG-only settings (R2, S3, S4) reach high sensitivity at low specificity, i.e., they label most windows as FoG.

\subsection{EEG encoder shaped by IMU, by leg movement}
\label{app:s4}
\begin{figure}[!h]
\centering
\includegraphics[width=\columnwidth]{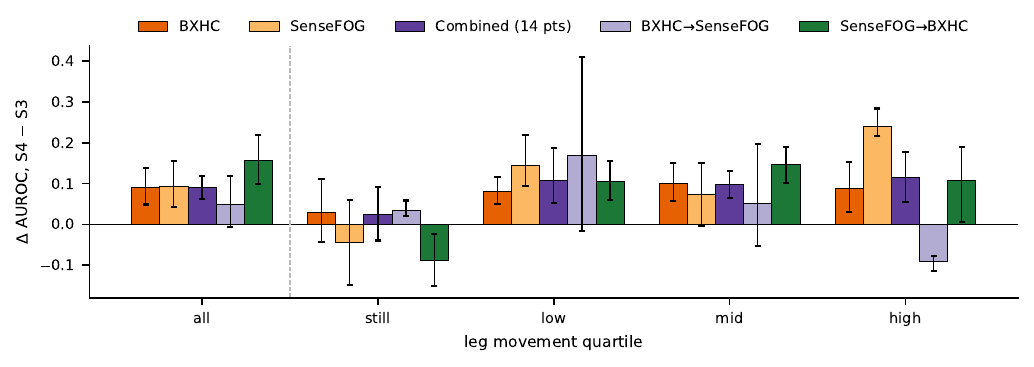}
\caption{AUROC change of S4 over S3 (EEG-only input) per leg-movement quartile, in all five evaluation settings. Mean over
patients with 95\% bootstrap confidence intervals.}
\label{fig:s4}
\end{figure}

Figure~\ref{fig:s4} compares the EEG encoder from EEG+IMU pretraining with the EEG-pretrained encoder, both with EEG-only input
(S4$-$S3), per leg-movement quartile in all five evaluation settings. In every setting, the gains are concentrated in the
three quartiles with leg movement (for example $+$0.08 to $+$0.10 on BXHC and $+$0.10 to $+$0.12 in the combined setting),
while the stillest quartile changes little or decreases ($-$0.09 to $+$0.04). The one exception is the high-movement quartile
under transfer from BXHC to SenseFOG ($-$0.09, three patients). Aligning EEG with IMU therefore improves EEG-only detection
mainly where the legs move, which is where EEG is most affected by movement-related signals~\cite{gwin2010removal}.

\begin{figure}[!h]
\centering
\includegraphics[width=\columnwidth]{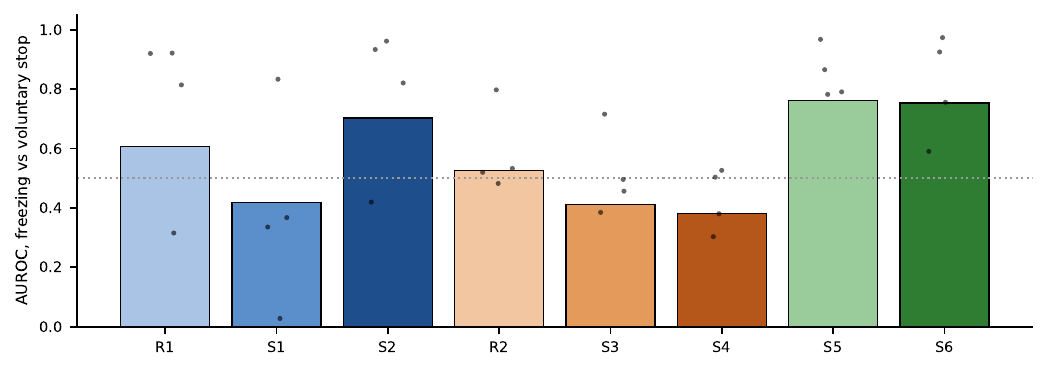}
\caption{SenseFOG: AUROC of the predicted FoG probability for freezing versus voluntary-stop windows. Bars: pooled over
windows (dominated by the patient with the most freezing windows); dots: patients.}
\label{fig:stops}
\end{figure}

\subsection{Freezing versus voluntary stops on SenseFOG}
\label{app:stops}
SenseFOG is the only dataset with annotated voluntary stops, which resemble freezing in that the legs are still. Figure~\ref{fig:stops}
reports, for each setting trained within SenseFOG, how well the predicted FoG probability separates freezing windows from
stop windows (AUROC; 489 freezing and 182 stop windows). The IMU-pretrained encoder separates them worse than chance (0.42) and
worse than the base encoder (0.61); its hyperparameters were selected on BXHC, which contains no annotated stops. The IMU
encoder from EEG+IMU pretraining (0.70) and both EEG+IMU-input settings (0.76 and 0.75) separate them best, while the EEG-only
settings stay near chance (0.38--0.53). With four patients, the per-patient values vary widely, so these results show direction
only.

\end{document}